\documentclass[preprints,article,submit,moreauthors,pdftex]{Definitions/mdpi}

\usepackage{times}
\usepackage{epsfig}
\usepackage{graphicx}
\usepackage{amsmath}
\usepackage{amssymb}

\usepackage{csvsimple}

\usepackage{booktabs}

\usepackage{algorithm,setspace}
\usepackage[noend]{algpseudocode}

\usepackage{xcolor}

\newcommand{\wa}[1]{{ #1}}

\preto{\abstractkeywords}{\nolinenumbers}

\firstpage{1} 
\pubvolume{xx}
\issuenum{1}
\articlenumber{5}
\pubyear{2019}
\copyrightyear{2019}
\history{Received: date; Accepted: date; Published: date}

\pdfoutput=1

\Title{Unsupervised and Generic Short-Term Anticipation of Human Body Motions}

\Author{Kristina Enes $^1$, Hassan Errami $^1$, Moritz Wolter $^2$, Tim Krake $^{3,4}$, Bernhard Eberhardt $^3$, Andreas Weber $^{1,}$, and J{\"o}rg Zimmermann $^1$*}

\AuthorNames{Kristina Enes, Hassan Errami, Moritz Wolter, Tim Krake, Bernhard Eberhardt, Andreas Weber,
Joerg Zimmermann}

\address{%
$^{1}$ \quad University of Bonn Germany, Visual Computing Department; weber@cs.uni-bonn.de\\
$^{2}$ \quad University of Bonn, Germany; wolter@cs.uni-bonn.de\\
$^{3}$ \quad HdM Stuttgart, Germany; eberhardt@hdm-stuttgart.de\\
$^{4}$ \quad University of Stuttgart; tim.krake@visus.uni-stuttgart.de
}

\corres{Correspondence: jz@cs.uni-bonn.de (J.Z.)}

\abstract{
Various neural network based methods are capable of anticipating human body motions from data for a short period of time. What these methods lack are the interpretability and explainability of the network and its results. 
We propose to use Dynamic Mode Decomposition with delays to represent and anticipate human body motions. Exploring the influence of the number of delays on the reconstruction and prediction of various motion classes, we show that the anticipation errors in our results are comparable or even better for very short anticipation times ($<0.4$ sec) to a recurrent neural network based method. We perceive our method as a first step towards the interpretability of the results by representing human body motions as linear combinations of ``factors''. In addition, compared to the neural network based methods large training times are not needed. Actually, our methods do not even regress to any other motions than the one to be anticipated and hence is of a generic nature.
}


\begin{document}

\maketitle

\section{Introduction}

Various kinds of neural network architectures are the main technical basis for the current state of the art for anticipation of human body motions from data
\cite{fragkiadaki2015recurrent,martinez2017human,gui2018adversarial,jain2016structural,li2018convolutional,pavllo2019modeling,abu2018will,ruiz2018human,gopalakrishnan2019neural}.
However, as is the case in many other application domains, there is a fundamental lack of interpretability of the neural networks.
In these approaches the two main conceptual ingredients of human motion prediction are also intermixed:
\begin{itemize}
    \item[(a)] Modelling the intent of the persons.
    \item[(b)] Modelling the influence of the previous motion.
\end{itemize}

Whereas for point (a) mechanistic models might be hard to obtain, for (b) models as dynamical systems partially reflecting bio-physical knowledge are possible in principle.
In this paper, we will focus on point (b).
Instead of suggesting another neural network based anticipation architecture, we will try to separate several possible constituents:
\begin{itemize}
\item Can recently developed so called equation free modeling techniques \cite{KutzBook16,SchmidSester08,Tu,fujii_supervised_2019,fujii_dynamic_2019,jovanovic_sparsity-promoting_2014} already explain and predict motions in a short time horizon? 

For this purpose the view of the motion time series as dynamical systems is used.
\item What is the role of incorporating delay inputs?
Many neural network architectures incorporate delays \cite{21701,Peddinti2015ATD,Huang_2019}, recurrent connections~\cite{hochreiter1997lstm, martinez2017human} or temporal convolutions~\cite{bai2018empirical, bai2018trellis, li2018convolutional}, but the contribution of the delays or the memory cannot be separated from the overall network architecture.
\end{itemize}

\wa{In general, dynamical systems are usually described via a set of differential equations.
For many systems a variety of appropriate data in form of observables are available.
However, if the process is complex the recovery of the underlying differential equation from data is a challenging task \cite{brunton_discovering_2016}.
Instead, the set of $m$ observables sampled at time steps $n$ is used for the investigation of the considered process.
For the identification of temporal structures, the Fourier theory is usually utilized.
Therefore, a Fourier analysis on the observables is performed, to extract amplitude and frequencies leading to a decomposition into trigonometric series.
This approach has some drawbacks for human motion capture data as these phenomena not exclusively consist of periodic components.
Hence, the decomposition will be distorted.
An algorithm that take this point into account is Dynamic Mode Decomposition (DMD).}

Introduced in the first version by Schmid and Sesterhenn in 2008 \cite{SchmidSester08}, DMD is a data-driven and---in its original form---``equation free'' algorithm extracting spatio-temporal patterns in the form of so-called DMD modes, DMD eigenvalues, and DMD amplitudes from the given observables. However, whereas Fourier-decomposition is obliged to use a fixed set of frequencies (i.e. eigenvalues), with DMD we detect and can use only the important ones and therefore adapt better to the process considered.

Since its introduction there has been many improvements and modification to this method.
One of these improvements is the combination of DMD with delay coordinates.
In this paper, we will investigate the use of DMD with delays (DMDd) in the context of short term anticipation of human motion from data.
\wa{Due to the structure of motion capture data (more time steps than observables) the use of delays optimizes the application of DMD.
Therefore, a more precise extraction of spatio-temporal patterns is achieved which leads an more accurate short-term prediction.}

Thus, the paper is structured as follow: In Section~\ref{sec:theo} we describe the DMD algorithm as well as Taken's Theorem and DMDd. In Section~\ref{sec:res} we explain our experiments with DMDd and give some examples of our results. In Section~\ref{sec:future} we discuss future possibilities given by our method.

\section{Theoretical background}
\label{sec:theo}

In this section, we clarify the background of Dynamic Mode Decomposition.
For the application on motion capture data we assume a vector-valued time series 
$x_1,x_2,\dots,x_n \in \mathbb{R}^m$, 
where each snapshot consists of \wa{marker positions (in 3D) or joint angles of a skeleton to a certain time step.}
Before we formulate the algorithm in more details, we briefly highlight the basic concept of DMD:
In a first step, the data were used to determine frequencies, the so-called DMD eigenvalues.
These are defined by the non-zero eigenvalues of a solution to the following minimization problem:
\begin{equation}
\min_{A \in \mathbb{C}^{m \times m}} \sum_{j=1}^{n} \lVert Ax_j - x_{j+1} \rVert_2^2.
\end{equation}
\wa{Then, the data were fitted to the previously computed frequencies (this process is similar to a discrete Fourier transformation or a trigonometric interpolation).}

However, in many application areas the number of observables is considerably larger than the number of snapshots, i.e. $m > n$.
Therefore, this approach leads to a sufficient number of frequencies and it can be proven that the reconstruction is error-free \cite{krake2019dynamic}.
For motion capture data, however, the converse is true, i.e. $m < n$.
Hence, in most cases we do not have enough frequencies for an adequate reconstruction, which even results in a bad anticipation as well. 

We approach this issue by manipulating the data in a preprocessing step, i.e. before applying EXDMD.
\wa{To this end, the theory of delays justified by Takens' Theorem is consulted, which is described in Subsection~\ref{ssec:TakensTheorem}.
Applying this technique leads to Dynamic Mode Decomposition with delay (DMDd) \cite{KutzBook16}.
The exact procedure is explained in Subsection~\ref{ssec:DMDD}.}

\subsection{Exact Dynamic Mode Decomposition}
\label{ssec:EXDMD}
EXDMD is the most modern variant of DMD that is applied directly on raw data.
It was published in 2014 by Tu et al. \cite{Tu}.
However, we have chosen the algorithmic formulation by Krake et al. \cite{krake2019dynamic}, which differs in the computation of DMD amplitudes.
Algorithm \ref{alg:EXDMD} shows an adjusted version of the algorithm. 
Since we mainly focus on anticipation, we are not interested in the reconstruction of the first snapshot and therefore some steps are skipped.

After defining the snapshot matrices $X$ and $Y$, which are related by one time-shift, a reduced singular value decomposition of $X$ is performed in line 2.
These components are used to determine the low-dimensional matrix $S$ that owns the dynamic relevant information in form of (DMD) eigenvalues $\lambda_j$.
Therefore, only the non-zero eigenvalues are used to compute the so-called DMD modes $\vartheta_j$ in line 7.
Finally, the DMD amplitudes are calculated via $a = \Lambda^{-1} \Theta^+ x_2$, where the second initial snapshot $x_2$ is used.

Given the DMD modes, DMD eigenvalues and DMD amplitudes we can both reconstruct the original snapshot matrix and make predictions for future states. But as mentioned before a good reconstruction might not be possible depending on the matrix dimensions. However if all conditions are met we can an exact reconstruction.

\begin{algorithm}
\setstretch{1.8}
\caption{Exact Dynamic Mode Decomposition}
	\label{alg:EXDMD}
	\begin{algorithmic}[1]
		\State Define $X = [ x_1 \; \dots \; x_{n-1} ]$, $Y = [ x_2 \; \dots \; x_{n}]$
		\State Calculate the reduced SVD $X = U \Sigma V^*$		
		\State Calculate $S = U^* Y V \Sigma^{-1}$ with $\textnormal{rank}(X) = r$

		\State Calculate $\lambda_1,\dots,\lambda_r$ and $v_1,\dots,v_r$ of $S$
		\For{$1 \leq i \leq r$}
		\If{$\lambda_i \neq 0$}
		\State $\vartheta_i = \frac{1}{\lambda_i} Y V \Sigma^{-1} v_i$
		\EndIf
		\EndFor
		\State $\Lambda = \text{diag}(\lambda_1,\lambda_2,\dots,\lambda_{r_0})$ with $\lambda_1, \lambda_2, \dots, \lambda_{r_0} \neq 0$
		\State $\Theta =  [\vartheta_1 \;  \vartheta_2 \; \dots \; \vartheta_{r_0} ]$
		\State Calculate $a = \Lambda^{-1} \Theta^+ x_2$ with $a = (a_1,\dots,a_{r_0})$
	\end{algorithmic}
\end{algorithm}

\subsection{Delay vectors and Takens' Theorem}
\label{ssec:TakensTheorem}
Most real world dynamical systems are only partially observable, i.e. we can observe only a low-dimensional projection of
a dynamical system acting on a high dimensional state space. This means that from a certain observed snapshot of a dynamical
system it is even in principle not possible to reconstruct the full current state of the dynamical system. Fortunately, the
information contained in observations made at several different time steps can be combined to reconstruct, at least in
principle, the complete current state, and (under certain technical assumptions) the dynamics on these delay vectors is
diffeomorphic to the true dynamics on the hidden state space. This delay embedding theorem is also known as Takens' theorem,
first proved by Floris Takens in 1981 \cite{Takens1981}. This result has led to a branch of dynamical systems theory now referred to as
``embedology'' \cite{Sauer1991}.

Here we give a brief sketch of the delay embedding theorem for discrete-time dynamical systems.
Let the state space of the dynamical system be a $k$-dimensional manifold $M$. The dynamics is defined by a smooth map
\begin{equation}
    \phi: M \rightarrow M,
\end{equation}
and the observations are generated by a twice-differentiable map $y: M \rightarrow \mathbb{R}$ (the observation function),
projecting the full state of the dynamical system to a scalar observable. From a time series of observed values we can
build $m$-dimensional \emph{delay vectors}:
\begin{equation}
    {\bold y_m(n)} = (y(n), y(n-1), y(n-m+1))^T.
\end{equation}
The delay vectors are elements of $\mathbb{R}^m$ and by mapping a delay vector to its successor we get a mapping $\rho$
from $\mathbb{R}^m$ to $\mathbb{R}^m$:
\begin{equation}
    \rho({\bold y_m(n)}) = {\bold y_m(n+1)}
\end{equation}

The delay embedding theorem now implies that the evolution of points ${\bold y_m(n)}$ in the reconstruction space $\mathbb{R}^m$
driven by $\rho$ follows (i.e., is diffeomorphic to) the unknown dynamics in the original state space $M$ driven by $\phi$ when $m = 2k+1$. Here $2k+1$
is a maximal value, faithful reconstruction could already occur for delay vectors of lower dimension.
Thus long enough delay vectors represent the full hidden state of the observed dynamical system, meaning that the
prediction of the next observed value based on a long enough history of past observations becomes possible.

For our purposes, we take the delay embedding theorem as an indication that adding delay dimensions to the observed state
vector can improve the anticipation quality of a DMD model.

\subsection{Dynamic Mode Decomposition with delays (DMDd)}
\label{ssec:DMDD}

Our motion capture data has the following form:
\begin{equation}
    X = \left[ 
        \begin{array}{ccc}
        x_1 & \hdots & x_n                                              
        \end{array}
        \right]
\end{equation}
Each state $x_i$ at time step $1\leqslant i \leqslant n$, is a vector of length $m$. To augment this matrix with $d$ delays we use a window of size $m \times (n-d)$, with $1<n-d<n$, to move along the motion data. This window starts at the first frame of $X$ and makes a copy of the first $n-d$ frames of the data, before taking a step of one frame along $X$. We continue with this process until the window reaches the end of the motion data. The copied data are then stacked one above the other resulting in a matrix $\tilde{X}$ with $n-d$ columns and $(d+1)m$ rows:
\begin{equation}
    \tilde{X} = \left[ 
        \begin{array}{ccc}
        x_1 & \hdots & x_{n-d}\\
        x_2 & \hdots & x_{n-d+1}\\
        \vdots & \ddots & \vdots\\
        x_{d+1} & \hdots & x_n
        \end{array}
        \right]
\end{equation}
Depending on how we choose $d$, the problem where our data has more columns than rows is no longer given. 
Applying the DMD algorithm to $\tilde{X}$ provides us with a good representation of the data and a good short-term future prediction is also possible, as will be detailed in Section \ref{sec:res}.

\section{Results}
\label{sec:res}
We tested DMDd on the Human3.6M dataset \cite{h36m_pami}, which consists of different kinds of actions like $walking, sitting$ and $eating$. These actions are performed by different actors. For our experiments we choose the motion sequences performed by actor number 5 (to have comparable results to the literature, as this actor was used for testing in the neural network based approaches, whereas the motions of the other actors were used for training). The data we use is sampled at 50\,Hz and contains absolute coordinates for each joint.
For each experiment we divide each action sequence into several sub-sequences of 100 or 150 frames length. The first 50 (1 sec) or 100 frames (2 sec) are taken as input for our methods and we compute a prediction for the next 5 frames (0.1 sec), 10 frames (0.2 sec) and 20 frames (0.4 sec).
To measure the distance between the ground truth $\mathit{GT}$ and our prediction $P$ we use two different distance measures. The first measure we use is the mean squared error (MSE):
\begin{equation}
    L(\mathit{GT},P) = \frac{1}{K} \sum_{k=1}^K \frac{1}{mp} \sum_{i,j} (\mathit{GT}_{ijk}-P_{ijk})^2
\end{equation}
$K$ is the number of motion sequences taken form the same action class and hence the number of predictions made for this action. Both $\mathit{GT}$ and $P$ consist of $m$ observables and $p$ frames.
The second distance measure we use is the Kullback-Leibler divergence as it was used in \cite{ruiz2018human}.

\subsection{Comparison with neural network based methods}

Before exploring the dependency of out methods with respect to various parameters we compare the setting of having the information of 1\,sec of motions as inputs (50 frames) using DMD with 80 delays with a RNN baseline as the one used in \cite{wolter-2018-nips}.
We use the mean squared error (MSE) as well as the Kullback-Leibler divergence as error measures for anticipation times of 0.1\,sec, 0.2\,sec, and  0.4\,sec.

The results given in Tables~\ref{tab:comparisonrnnrmse} and  \ref{tab:comparisonrnnkl1} indicate that our method shows better results for 0.1\,sec, 0.2\,sec and for most motion classes even for 0.4\,s, although they are not only unsupervised but even no knowledge about any other motion is taken into account! Interestingly, the error of the RNN slightly decreases with the anticipation times. This counter-intuitive behavior of the RNN approach might be explained by the fact that the anticipations yielded by the RNN baseline in general shows small jumps at the beginning of the anticipation period \cite{mao2019learning}.

\begin{table}[ht]
    \centering
\begin{filecontents*}{Tables/50inrnndmdd80mse.csv}
Action,RNN 0.1 sec,RNN 0.2 sec,RNN 0.4 sec,DMDd 0.1 sec,DMDd 0.2 sec,DMDd 0.4 sec
Directions,787.18,687.42,1202.92,9.09,103.03,1097.34
Discussion,964.55,874.03,1536.45,26.10,187.37,1459.77
Eating,626.26,490.74,599.55,13.16,119.89,1250.54
Greeting,1316.82,1337.01,2748.43,46.86,471.66,4096.82
Phoning,832.94,699.94,945.87,15.60,150.46,6755.74
Posing,1184.45,1118.47,2067.22,16.12,193.23,1981.44
Purchases,1176.26,1069.58,1905.41,74.84,510.89,5717.04
Sitting,1008.65,917.76,1390.35,14.05,86.67,665.49
SittingDown,11164.34,8584.16,8784.89,207.44,750.49,4183.22
Smoking,778.76,610.34,754.22,14.37,95.79,789.76
Photo,2705.89,2070.79,2576.26,15.09,113.18,1038.58
Waiting,1081.52,886.59,1314.19,18.77,169.84,1862.09
Walking,821.12,635.70,776.37,53.82,397.00,3054.54
WalkDog,3083.24,2730.66,4248.07,91.59,588.56,3851.82
WalkTogether,1163.97,919.58,1198.45,15.00,144.91,1306.06
\end{filecontents*}
\csvautotabular{Tables/50inrnndmdd80mse.csv}
    \caption{Comparison of anticipation error using a RNN and Dynamic Mode Decomposition with 80 delays for various anticipation times. The error measure is the mean squared error on the pose sequences for anticipation times of 0.1\,sec (5 frames), 0.2\,sec (10 frames), and  0.4\,sec (20 frames) on the different motion classes of the Human 3.6M dataset for actor \#5.}
    \label{tab:comparisonrnnrmse}
\end{table}

\begin{table}[ht]
    \centering
\begin{filecontents*}{Tables/50inrnndmdd80kl1.csv}
Action,RNN 0.1 sec,RNN 0.2 sec,RNN 0.4 sec,DMDd 0.1 sec,DMDd 0.2 sec,DMDd 0.4 sec
Directions,0.38,0.18,0.07,0.02,0.02,0.03
Discussion,0.40,0.13,0.07,0.01,0.02,0.03
Eating,0.37,0.15,0.06,0.03,0.04,0.07
Greeting,0.53,0.17,0.10,0.04,0.03,0.04
Phoning,0.26,0.16,0.07,0.02,0.03,0.03
Posing,0.50,0.19,0.10,0.02,0.03,0.03
Purchases,0.32,0.14,0.08,0.01,0.02,0.02
Sitting,0.57,0.28,0.10,0.02,0.02,0.04
SittingDown,0.74,0.35,0.17,0.01,0.02,0.03
Smoking,0.45,0.14,0.06,0.02,0.02,0.02
Photo,0.54,0.27,0.12,0.02,0.02,0.03
Waiting,0.28,0.12,0.05,0.01,0.03,0.03
Walking,0.27,0.12,0.06,0.03,0.03,0.03
WalkDog,0.48,0.19,0.09,0.03,0.03,0.04
WalkTogether,0.48,0.18,0.08,0.03,0.04,0.04
\end{filecontents*}
\csvautotabular{Tables/50inrnndmdd80kl1.csv}
    \caption{Comparison of anticipation error using a RNN and Dynamic Mode Decomposition with 80 delays for various anticipation times. The error measure is the Kullback-Leibler divergence for anticipation times of 0.1\,sec (5 frames), 0.2\,sec (10 frames), and  0.4\,sec (20 frames) on the different motion classes of the Human 3.6M dataset for actor \#5.}
    \label{tab:comparisonrnnkl1}
\end{table}

\subsection{Reconstruction and anticipation of motions using DMD and DMD with delays}

Adding time delays already improves the reconstructibility of motions. In Table~\ref{tab:comparisonrmse} we show the average reconstruction errors 
of motion clips of 2\,sec length (100 frames) for the different motion classes. Already adding 10 time delays yields a dramatic improvement.  After adding 60 delays the reconstruction error drops to less than $10^{-5}$ for all motion classes.

\begin{table}[ht]
    \centering
\begin{filecontents*}{Tables/recondmddifdelaysmse.csv}
Action,DMD,DMDd10,DMDd20,DMDd30,DMDd40,DMDd50,DMDd60
Directions,3.20E+13,1.11E+01,5.99E-03,2.56E-05,2.22E-05,1.15E-05,7.61E-06
Discussion,2.20E+40,5.02E-04,7.93E-05,3.76E-05,3.39E-05,1.51E-05,9.89E-06
Eating,1.61E+08,1.07E-04,5.54E-05,3.42E-05,1.90E-05,1.21E-05,7.84E-06
Greeting,2.77E+07,8.85E-04,9.43E-05,3.57E-05,2.14E-05,1.07E-05,7.80E-06
Phoning,1.70E+18,7.55E+09,1.74E+08,1.01E+03,3.30E-03,1.73E-05,7.46E-06
Posing,1.00E+50,9.75E+03,5.03E-04,9.42E-05,1.71E-05,1.42E-05,8.38E-06
Purchases,7.94E+44,7.30E-04,6.63E-05,2.52E-05,2.58E-05,1.70E-05,6.78E-06
Sitting,3.45E+27,9.67E+07,2.15E+08,1.95E+06,5.61E+00,1.57E-05,3.46E-06
SittingDown,1.09E+45,4.64E+05,1.19E-04,1.41E-05,1.11E-05,6.51E-06,3.34E-06
Smoking,1.32E+20,2.45E+00,1.95E-02,1.15E-04,2.18E-05,1.18E-05,6.70E-06
Photo,7.66E+24,1.27E+00,6.56E-03,3.47E-05,4.04E-05,2.16E-05,8.31E-06
Waiting,3.27E+27,6.59E+01,5.86E-01,2.26E-04,6.55E-05,1.49E-05,4.43E-06
Walking,5.12E+13,1.07E-04,7.83E-05,3.57E-05,3.05E-05,2.51E-05,1.51E-05
WalkDog,3.27E+19,2.37E+05,1.68E-04,5.02E-05,2.47E-05,2.42E-05,1.07E-05
WalkTogether,7.34E+06,7.36E-05,7.03E-05,1.87E-05,3.30E-05,2.28E-05,1.18E-05
\end{filecontents*}
\csvautotabular{Tables/recondmddifdelaysmse.csv}
    \caption{Comparison of reconstructions errors using Dynamic Mode Decomposition (DMD)  and Dynamic Mode Decomposition with delays for various numbers of delays (10, 20, 30, 40, 50, and 60). The error measure is the mean squared error on the pose sequences of length 2\,sec on the different motion classes of the Human 3.6M dataset for actor \#5.}
    \label{tab:comparisonrmse}
\end{table}

The results of the anticipation errors  for 0.4\,sec (20 frames) of anticipation using 2\,sec  (100 frames) as context length is given in 
Table \ref{tab:antipdiffdelays}. The anticipation errors for DMD without delays is large (>$10^{10}$ for all motion classes and is not reproduced in the Table. 
In contrast to the reconstruction case, in which the error monotonically decreases with adding additional delays, the anticipation errors have minima at a certain number of delays (ranging between 40 and 90 for the different motion classes.

\begin{table}[ht]
    \centering
\begin{filecontents*}{Tables/antipdmddifdelaysmse.csv}
Action,DMDd10,DMDd20,DMDd40,DMDd50,DMDd60,DMDd70,DMDd80,DMDd90
Directions,1.87E+04,1.84E+03,1.03E+03,1.13E+03,9.29E+02,8.46E+02,8.39E+02,8.45E+02
Discussion,2.30E+03,1.47E+03,1.38E+03,1.31E+03,1.24E+03,1.23E+03,1.19E+03,1.16E+03
Eating,1.62E+03,8.71E+02,6.24E+02,7.48E+02,7.29E+02,7.23E+02,7.11E+02,6.93E+02
Greeting,4.56E+03,2.86E+03,2.53E+03,2.64E+03,2.51E+03,2.20E+03,1.99E+03,1.99E+03
Phoning,2.13E+13,2.30E+13,1.87E+03,1.25E+03,9.71E+02,8.59E+02,8.12E+02,8.29E+02
Posing,2.71E+08,2.14E+03,1.90E+03,1.45E+03,1.20E+03,1.06E+03,1.06E+03,1.06E+03
Purchases,2.64E+03,3.28E+03,2.79E+03,2.22E+03,2.09E+03,1.51E+03,1.35E+03,1.30E+03
Sitting,6.07E+12,5.31E+13,2.43E+03,8.01E+02,5.88E+02,5.81E+02,5.69E+02,5.64E+02
SittingDown,1.64E+10,1.85E+03,1.42E+03,1.31E+03,1.14E+03,1.11E+03,1.09E+03,1.13E+03
Smoking,3.83E+03,9.47E+02,8.08E+02,8.30E+02,8.04E+02,7.49E+02,7.65E+02,7.21E+02
Photo,1.00E+04,2.00E+03,1.50E+03,1.48E+03,1.35E+03,1.22E+03,1.16E+03,1.11E+03
Waiting,1.90E+06,1.30E+04,1.27E+03,1.18E+03,1.20E+03,1.19E+03,1.19E+03,1.18E+03
Walking,2.99E+03,1.72E+03,1.70E+03,1.87E+03,2.06E+03,2.05E+03,2.07E+03,2.12E+03
WalkDog,8.73E+10,4.14E+03,3.95E+03,3.68E+03,3.82E+03,3.80E+03,4.05E+03,4.75E+03
WalkTogether,1.35E+03,9.91E+02,9.93E+02,1.04E+03,1.09E+03,1.11E+03,1.08E+03,1.12E+03
\end{filecontents*}
\csvautotabular{Tables/antipdmddifdelaysmse.csv}
    \caption{Comparison of anticipation  errors for  0.4\,sec (20 frames) using Dynamic Mode Decomposition with delays for various numbers of delays (10, 20, 40, 50, 60, 70, 80 and 90). The error measure is the mean squared error on the pose sequences of length 2\,sec on the different motion classes of the Human 3.6M dataset for actor \#5.}
    \label{tab:antipdiffdelays}
\end{table}

\subsection{Using different input lengths of motions to be anticipated}

\begin{figure}[ht]
    \centering
\includegraphics[width=0.7\textwidth]{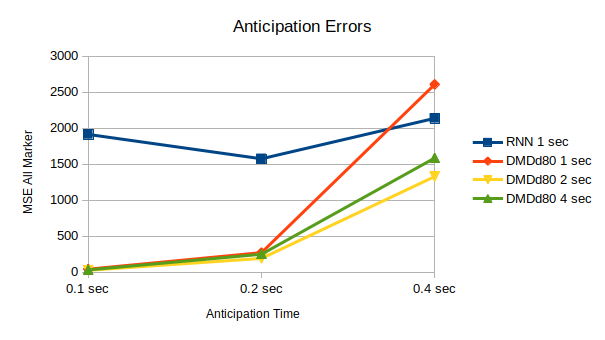}
    \caption{Comparison of anticipation  errors for  anticipations of  0.1\,sec (5 frames),  0.2\,sec (10 frames), 
    and 0.4\,sec (20 frames). The result of a trained RNN using inputs of 1\,sec, and DMDd80 on inputs of 1\,sec, 2\,sec, and 4\,sec.
    The error measure is the average of the  mean squared error on the pose sequences  on the different motion classes of the Human 3.6M dataset for actor \#5.}
    \label{fig:compantip124}
\end{figure}

We compare the previously used setting of having the information of 1\,sec of motions as inputs (50 frames) to the one with 2\,sec of motions as inputs (100 frames), and
4\,sec of motions as inputs (200 frames).
In Fig.~\ref{fig:compantip124} we show the MSE for the anticipation of a trained RNN with 1\,sec of motions as inputs, the DMDd with 1\,sec of motions as inputs, DMDd with 2\,sec of motions as inputs, and  DMDd with 4\,sec of motions as input (for anticipation times of 0.1\,sec, 0.2\,sec, and 0.40\,sec).

\subsection{Reconstruction and anticipation of inertial measurements of root and end-effectors}

For assessing short term anticipation on the basis of sparse accelerometer data attached 
to the end effectors and the hip 
we used also the marker position data of the Human 3.6M database to have a 
large collection of motions and ``ground truth data''.
As has already been shown in \cite{tautges-2011-MotionReconstruction} the use the second time 
derivatives of  marker position data yields reliable estimates for tests using data of accelerometers.

\begin{figure}[ht]
    \centering
    \begin{tabular}{cc}
\includegraphics[width=0.48\textwidth]{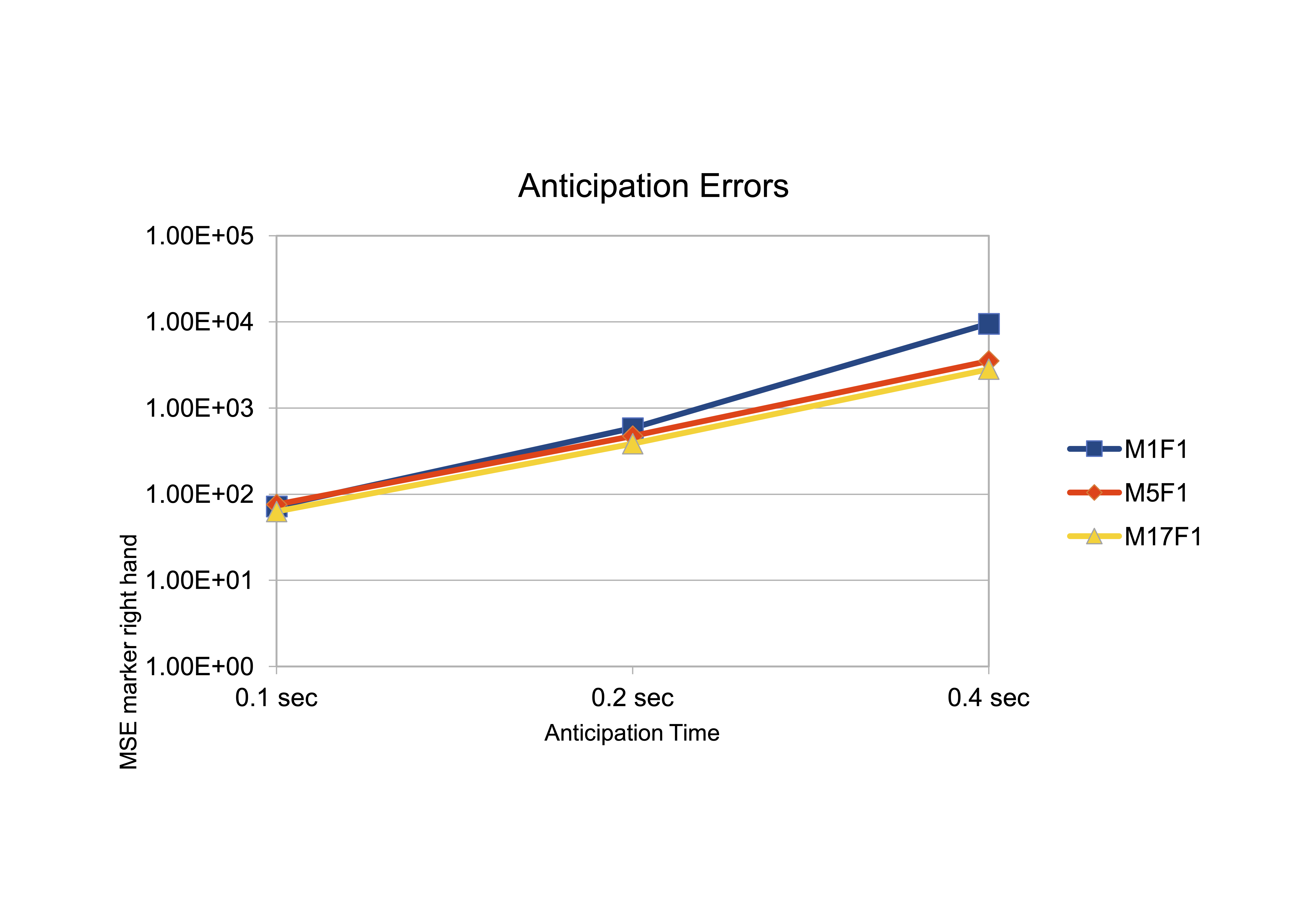}&
\includegraphics[width=0.48\textwidth]{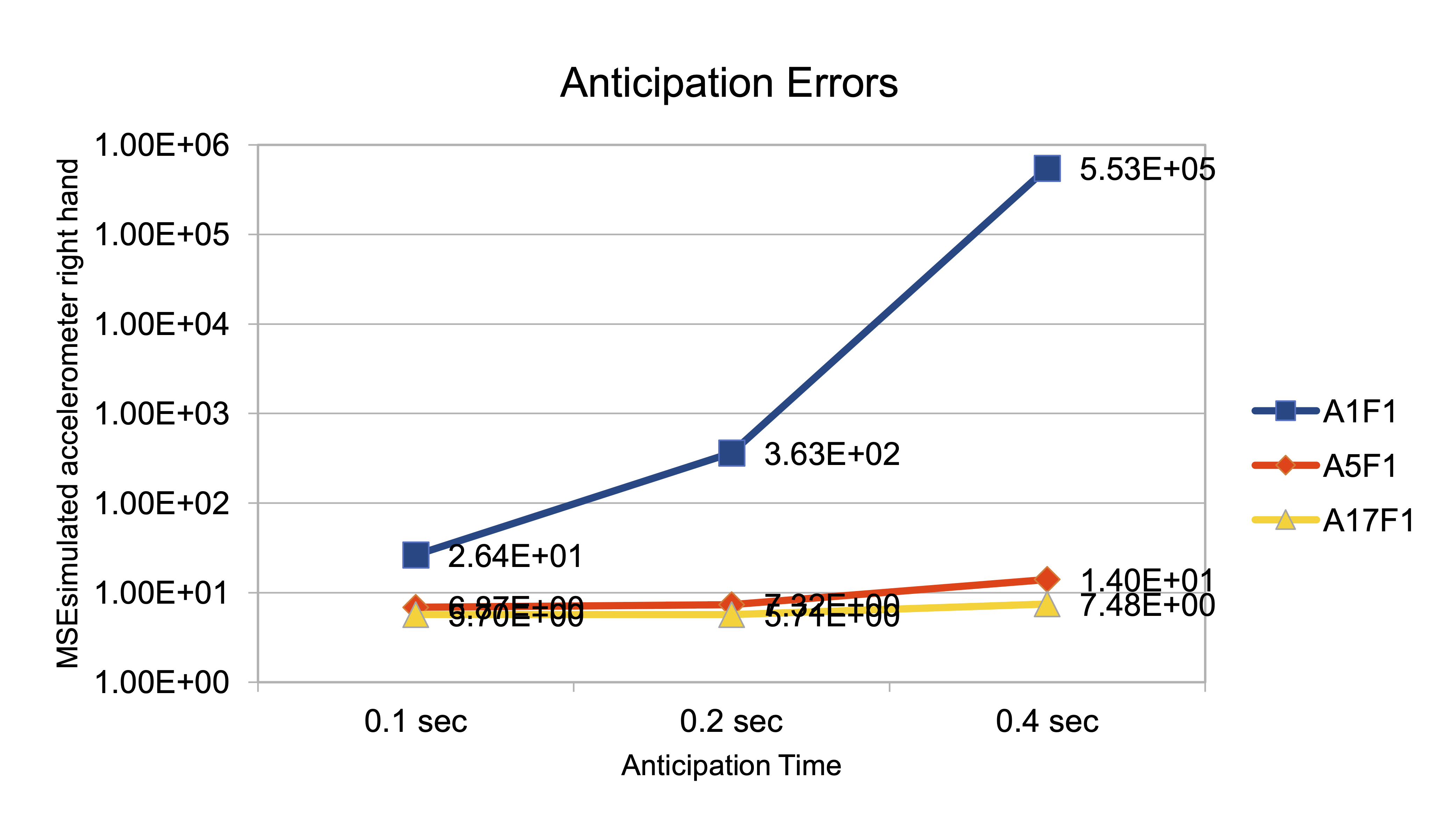}
\end{tabular}
    \caption{Left: Single marker anticipation without and with spatial context. The anticipation errors of  0.1\,sec (5 frames),  0.2\,sec (10 frames), 
    and 0.4\,sec (20 frames) are given for the anticipation of marker of the right hand joint using no spatial context (M1F1), all end effector markers and the root position as spatial context (M5F1), and all 17 markers as spatial context (M17F1).
    The error measure is the average of the  mean squared error on the marker sequences of the right hand joint on the different motion classes of the Human 3.6M dataset for actor \#5.
    Right: Same as left, but for second derivatives of marker positions simulating accelerometer sensors. Notice that the error is given in logarithmic scale.}
    \label{fig:1marker}
\end{figure}

In Figure \ref{fig:1marker} the results of the anticipation error of just the marker of the right hand is given. The anticipation error of performing the DMDd80 on the time series of just this one marker is given as M1F1. The error of this one marker but using DMDd80 on the end effectors and the root is given as M5F1; the one performing DMDd80 on all 17 markers is given as M17F1. Using second derivatives as simulation of accelerometer sensor data are given similarly as A1F1, A5F1, and A17F1.
Whereas the addition of a ``spatial context'' of other markers than the one measured for anticipation in the DMDd computation has little effect for 0.1\,sec  of anticipation time, there is a considerable effect for 0.2\,sec  of anticipation time, and a huge effect for 0.4\,sec of anticipation time: For the simulated accelorometers the average MSE  of the right hand marker with a value of 5.53E+05 was about four orders of magnitude larger when performing the DMDd only on its time series compared to the one when using the spatial context of the 4 additional ones (at left hand, left and right foot, and root) with a value of 1.40E+01!
Considering more than 4 additional markers had little additional effect.

\section{Conclusion and Future Work}
\label{sec:future}

In contrast to some special classes of human motions, on which the direct application of DMD to the observables of human motion data
can be suitable for a good  reconstruction of the data \cite{takeishi_bayesian_2017,fujii_supervised_2019}, this direct applications of DMD to the observables of the
 motions contained in the Human 3.6M dataset do not yield  good reconstructions, nor suitable short-term anticipations.

Inspired by Takens' theorem, which
emphasizes the usefulness of delays in reconstructing partially
observable, high dimensional dynamics, we have extended DMD
with delay vectors of different length and evaluated
the impact on short-term anticipation using a large
real world human motion data base and comparing the performance
to a state of the art RNN model.
The results show that delays can drastically improve reconstruction and also
anticipation performance, often by several orders of magnitude, and
in many cases lead to better anticipation performance than
the RNN model (for anticipation times less than 0.4\,sec). This is especially remarkable, as our methods do not even regress to any other motions than the one to be anticipated. Moreover, DMD effectively
solves a convex optimization problem and thus is
much faster to evaluate than training RNNs.
Additionally, solutions of convex optimization problems are
globally optimal, a guarantee which is absent for trained RNNs.

As already mentioned in the introduction the presented work was primarily concerned with modelling the influence of the previous motion on the motion anticipation. For modeling the intent of persons other methods are required, and neural network based methods might be the ones of choice. Coming up with a hybrid DMD and neural network based method for mid-term (or even long-term) motion anticipation will be the topic of future research.

As a final remark, we mention that linear methods like DMD can foster the interpretability
of results by representing the evolution of motion as a linear
combination of ``factors'', where factors can be previous states,
delays, or nonlinear features computed from
the previous states or delays. This could prove to be especially useful when
machine learning driven systems enter more and more critical application areas,
involving aspects of security, safety, privacy, ethics, and politics. To address
these concerns, and for many application areas involving anticipation of human motions
these concerns play a central role, transparency, explainability, and interpretability
become more and more important criteria for the certification of machine learning
driven systems. For a comprehensive review of the current literature addressing
these rising concerns about safety and trustworthiness in machine learning see \cite{Hutter18}.

\authorcontributions{Conceptualization, A.W., J.Z., and B.E.; methodology, A.W, J.Z., T.K., B.E, and M.W. ; software, K.E., M.W., H.E.; validation, K.E., H.E., M.W.,  and J.Z.; formal analysis, J.Z., T.K., and M.W; data curation, M.W., K.E., H.E.; writing--original draft preparation, all; writing--review and editing, all; visualization, K.E., H.E., M.W.}

\funding{This work has been supported in part by {\em Deutsche Forschungsgemeinschaft} under grant 313421352 (DFG-Forschergruppe 2535 ``Anticipating Human Behavior'', projects P3 and P4).
}

\conflictsofinterest{The authors declare no conflict of interest.} 

\abbreviations{The following abbreviations are used in this manuscript:\\

\noindent 
\begin{tabular}{@{}ll}
DMD  & Dynamic Mode Decomposition\\
DMDd  & Dynamic Mode Decomposition with delays\\
DMDd$n$  & Dynamic Mode Decomposition using $n$ delays\\
RNN & Recurrent Neural Network\\
MSE & Mean squared error\\
\end{tabular}}

====================================
\externalbibliography{yes}
\bibliography{dmdmotion}

\end{document}